\UseRawInputEncoding
\documentclass{article}

\pdfoutput=1

\usepackage[nonatbib, preprint]{neurips_2023}

\usepackage[utf8]{inputenc} 
\usepackage[T1]{fontenc}    
\usepackage{hyperref}       
\usepackage{url}            
\usepackage{booktabs}       
\usepackage{amsfonts}       
\usepackage{nicefrac}       
\usepackage{microtype}      
\usepackage{xcolor}         
\usepackage{graphicx}
\usepackage{ulem}
\usepackage{multirow}

 \title{Fashion Matrix: Editing Photos by Just Talking}

\author{
  Zheng Chong$^{1,2}$   \quad  Xujie Zhang$^{1}$   \quad  Fuwei Zhao$^{1}$ \quad 
  Zhenyu Xie$^{1}$  \quad    Xiaodan Liang$^{1,2}$\thanks{Corresponding author is Xiaodan Liang (\texttt{xdliang328@gmail.com}).}   \\
  $^1$Shenzhen Campus of Sun Yat-Sen University  \\
  $^2$Peng Cheng Laboratory\\
  \url{https://zheng-chong.github.io/FashionMatrix} \\
}

\newif\ifcomments
\commentsfalse 

\begin{document}

\maketitle
\begin{center}
    \captionsetup{type=figure}
    \includegraphics[width=1\textwidth]{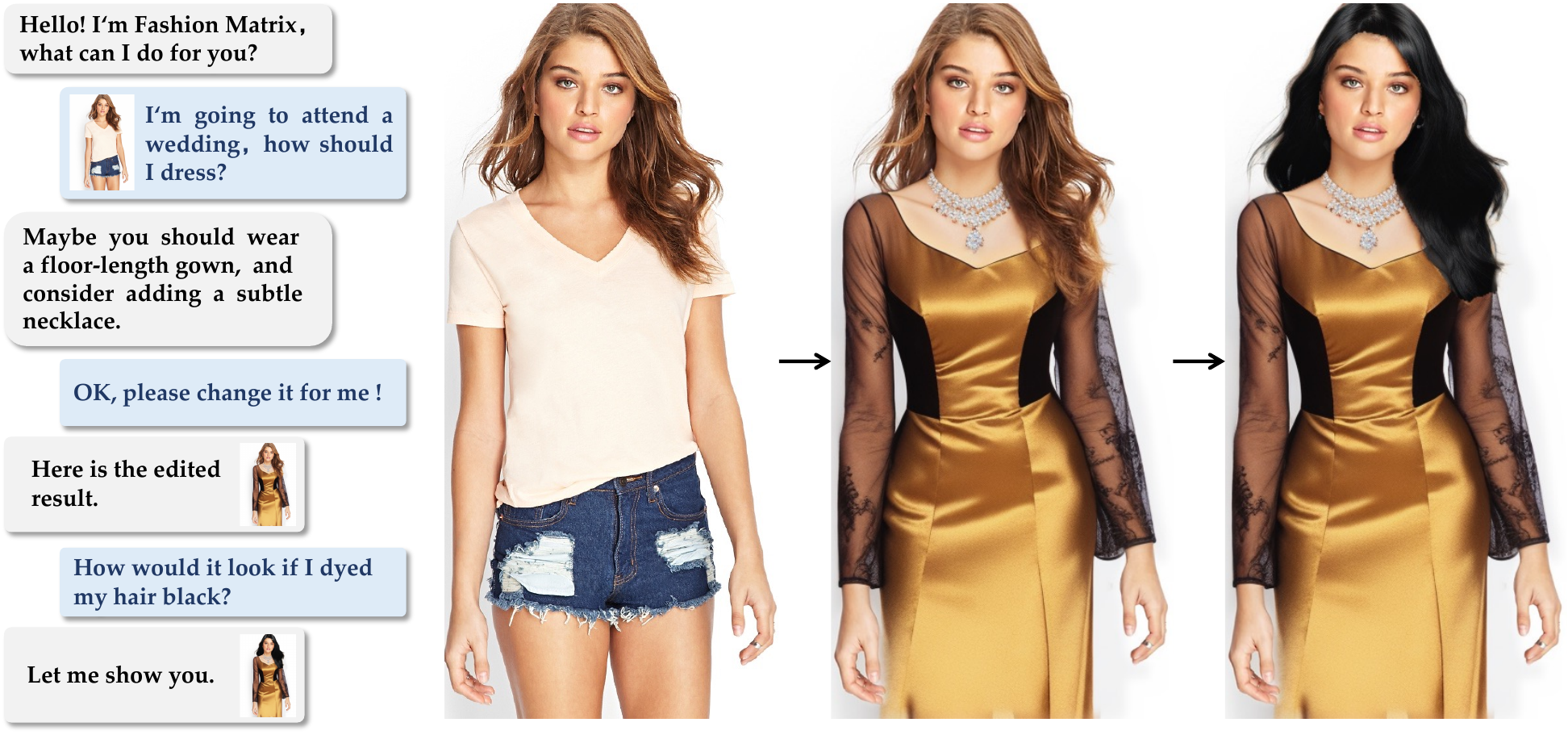}
    \captionof{figure}{Fashion Matrix demonstrates the capacity for engaging in multiple rounds of user dialogue, enabling proficient and precise photo editing of individuals based on provided instructions.}
\end{center}


\begin{abstract}
The utilization of Large Language Models (LLMs) for the construction of AI systems has garnered significant attention across diverse fields. The extension of LLMs to the domain of fashion holds substantial commercial potential but also inherent challenges due to the intricate semantic interactions in fashion-related generation.
To address this issue, we developed a hierarchical AI system called \textbf{Fashion Matrix} dedicated to editing photos by just talking. This system facilitates diverse prompt-driven tasks, encompassing garment or accessory replacement, recoloring, addition, and removal.
Specifically, Fashion Matrix employs LLM as its foundational support and engages in iterative interactions with users. It employs a range of Semantic Segmentation Models (e.g., Grounded-SAM, MattingAnything, etc.) to delineate the specific editing masks based on user instructions. Subsequently, Visual Foundation Models (e.g., Stable Diffusion, ControlNet, etc.) are leveraged to generate edited images from text prompts and masks, thereby facilitating the automation of fashion editing processes.
Experiments demonstrate the outstanding ability of Fashion Matrix to explores the collaborative potential of functionally diverse pre-trained models in the domain of fashion editing.
The code is available at \url{https://github.com/Zheng-Chong/FashionMatric}.

\end{abstract}

\begin{figure}
  \centering
  \includegraphics[width=1.0\hsize]{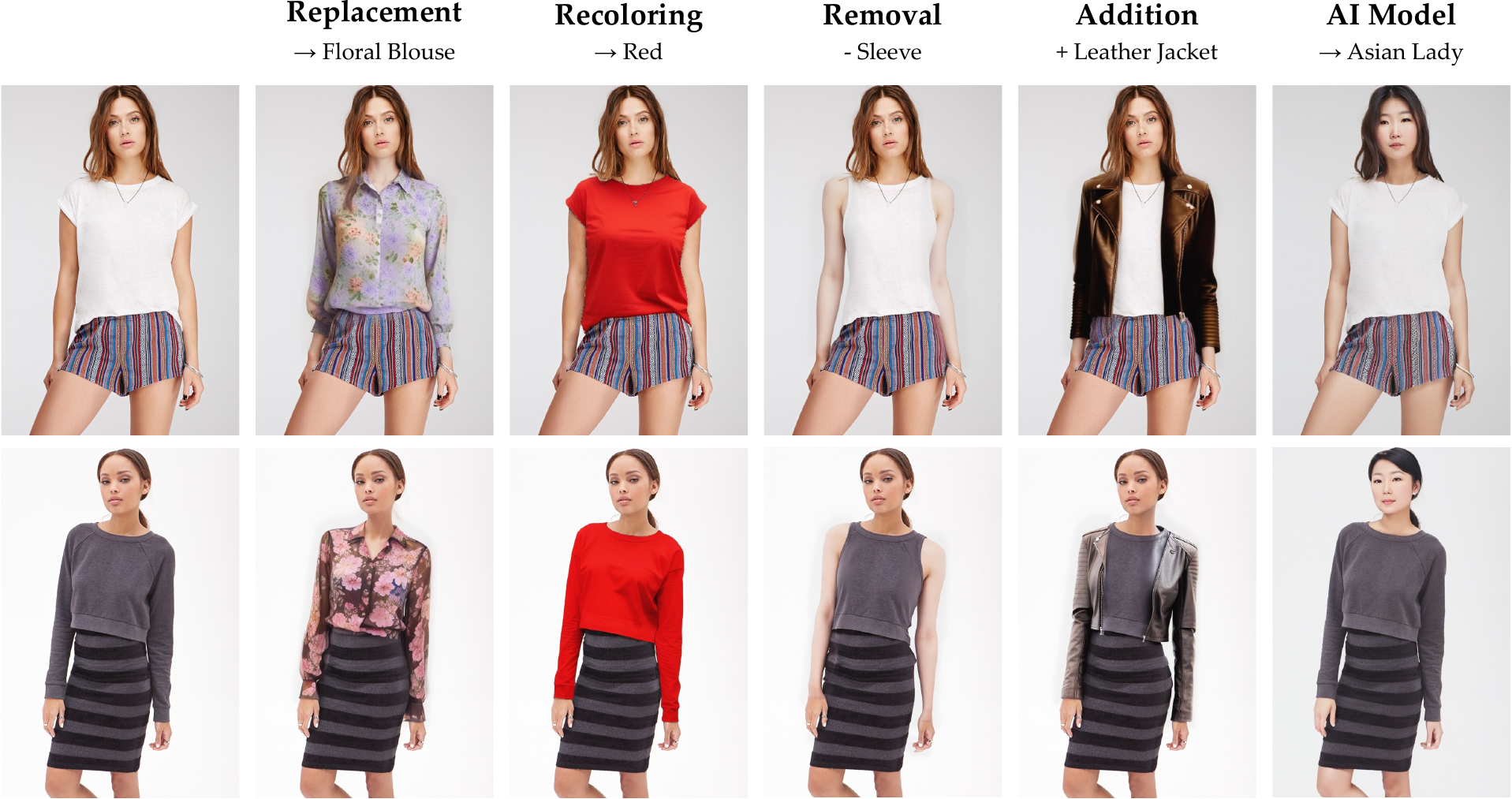}
  \caption{Fashion Matrix can perform multi-functional fine-grained fashion editing based on provided instructions while ensuring that the original image information is conserved to the greatest extent possible.}
  \label{fig:teaser}
\end{figure}

\section{Introduction}


Recently, large language models (LLMs) such as PaLM\cite{chowdhery2022palm}, LLaMA\cite{touvron2023llama}, and ChatGPT\cite{InstructGPT} have proven highly effective for various Natural Language Processing (NLP) tasks, such as knowledge graphs, code completion, and chat robots, etc.
Moreover, innovators begin extending LLMs to domain-specific tasks and producing agent systems that demonstrate remarkable proficiency in handling complex problems. These works typically endow LLMs with the functionalities of other models or tools, thereby augmenting their capacity to address diverse application scenarios beyond NLP. 

For instance, Visual ChatGPT\cite{wu2023visual} integrates a spectrum of Visual Foundation Models, empowering LLMs with adeptness in reading, editing, and reconstructing images.
Visual ChatGPT exhibits satisfactory performance in general-purpose neutral image editing. However, for the fashion-related domain, which focuses on human-centric generation and editing, Visual ChatGPT obtains inferior performance due to the lack of dedicated semantic perception for human body (e.g., human pose, human parsing, etc.).
This highlights the need for the combination of cutting-edge image generation model with the advanced semantic modeling model to further facilitate the improvement of LLM for fashion-related application scenario.

To tackle this concern, we have developed a multi-round dialogue AI system named \textbf{Fashion Matrix}, expressly tailored for fashion-centric applications. Serving as the pioneer in conversational fashion editing, this innovative framework integrates LLMs with cutting-edge image generation models (e.g. Stable Diffusion\cite{rombach2021highresolution}, ControlNet\cite{zhang2023Controlnet}, etc.) and semantic segmentation models (e.g., Grounded-SAM\cite{kirillov2023segany,liu2023grounding}, MattingAnything\cite{li2023matting}, etc.) facilitating expeditious and accurate guidance for multiple editing tasks (as shown in Fig. \ref{fig:teaser}).

\begin{figure}
  \centering
  \includegraphics[width=1\hsize]{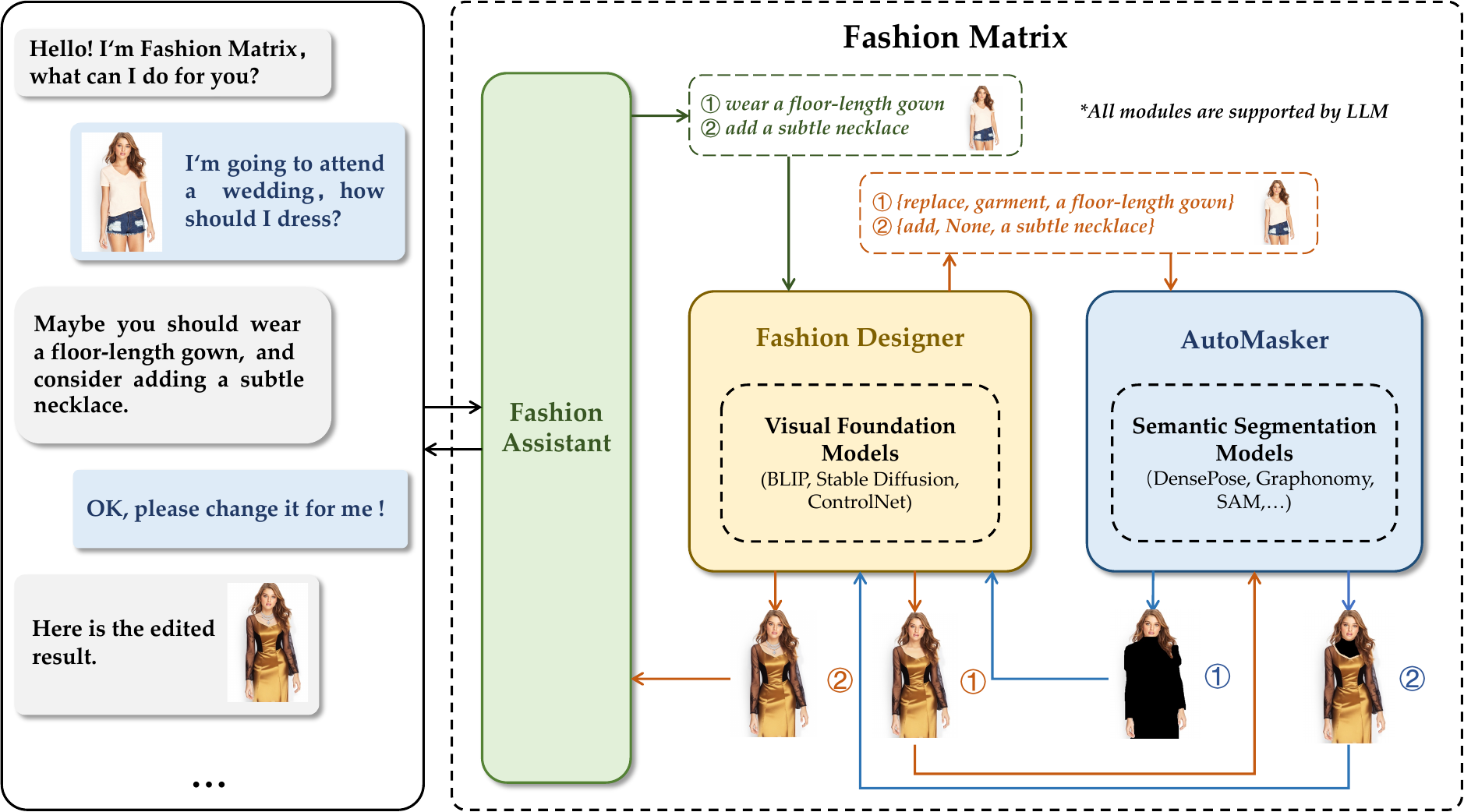}
  \caption{\textbf{Overview of the system hierarchy}. The system is composed of three modules: (1) \textit{Fashion Assistant} , (2) \textit{Fashion Designer}, and (3) \textit{AutoMasker}, which are at different levels, and all of them use LLM as the support of intelligent text processing. \textit{Fashion Assistant} engages in user interactions to collect requirements, which are subsequently examined and transformed into instructions by \textit{Fashion Designer}. \textit{AutoMasker} identifies the editing region based on the semantic context of the instructions. Hierarchical design simplifies the logical processing flow and facilitates efficient information processing.}
  \label{fig:system}
\end{figure}

Specifically, our Fashion Matrix is composed of three modules as shown in Fig.~\ref{fig:system}: (1) \textit{Fashion Assistant} , (2) \textit{Fashion Designer}, and (3) \textit{AutoMasker}. 
The Fashion Assistant employs LLM to engage in dialogues with users, and gathers their editing requirements.
Fashion Designer designed for logic control plays a core role in the whole system. It partitions the user's editing requirements into discrete editing tasks, followed by prompt standardization for each task and utilizes AutoMasker and the corresponding visual fondation models to iteratively process the tasks.
AutoMasker is crucial for achieving fine-grained and open-vocabulary editing capabilities. It combines results form multiple semantic segmentation models 

To summarize, we present three main contributions:
\begin{itemize}
    \item We propose Fashion Matrix, a conversational system with a structured hierarchical architecture. It can address diverse fashion editing tasks bolstered by the integration of LLM,  Semantic Segmentation Models and Visual Foundation Models 
    \item We propose an \textit{AutoMasker} module that integrates various human parsing, pose estimation and general semantic segmentation models to from a new fine-grained human segmentation map named CoSegmentation and generate task-oriented semantic masks, facilitating a wide range of fashion editing tasks.
    \item Extensive zero-shot experiments have demonstrated the exceptional performance of our Fashion Matrix. Its versatility makes it valuable for both professional designers and casual users who wish to explore various outfit combinations and styles.
\end{itemize}

\section{Related Work}

\subsection{Agent System}

AutoGPT\cite{richards2023auto}, GPT-Engineer\cite{gpt-engineer}, HuggingGPT\cite{shen2023hugginggpt}, BabyAG\cite{babyagi3}, and other projects have demonstrated to a certain extent the ability to use a large language models (LLMs) as the core controller to build an Agent System. The potential of LLMs is not limited to generating content, stories, papers, etc. It also has powerful general problem-solving capabilities that can be applied in various fields.
In the LLM-driven AI Agent System, LLM is the "brain" of the system, which uses Chain-of-Thought (CoT)\cite{li2020manigan-cot, zhang2022automatic-cot}, ReAct\cite{yao2022react} , and other ways to think about the specified target person and obtain the target result by calling external tools. Although there are plenty of studies on Agent Systems, the incorporation of LLM capabilities into fashion-related domains remains a relatively underexplored area.


\subsection{Human Parsing and Pose Estimation}

Human parsing and pose estimation belong to the human-centered subdivision of dense prediction tasks, which supports the development of virtual try-on and fashion-related generation. OpenPose\cite{openpose}, MMPose\cite{mmpose2020}, and other methods\cite{alphapose, li2019crowdpose} identify the specified keypoints of the human body in the picture and form a pose heatmap in the form of a skeleton. DensePose\cite{Guler2018DensePose} realizes the mapping of 2D RGB images to 3D models, which has richer information than skeleton, and its prediction segmentation map also has clothing-agnostic features. Graphonomy\cite{Gong2019Graphonomy} and some other works\cite{gong2018instance-cihp, Gong_2017_CVPR-ssl, liang2018look-JPPnet} can identify and segment parts with specified semantics (such as top, coats, hair, etc.), but their segmentation is limited to specified labels, making it difficult to perform finer-grained division.
Recently, SAM\cite{kirillov2023segany} achieves open-domain segmentation when providing prompts (such as boxes / points), which is a landmark progress in the field of dense prediction. Grounded-SAM achieves open-domain segmentation through text prompts by combining GroundingDINO\cite{liu2023grounding} and SAM\cite{kirillov2023segany} without manually labeling the bounding box. Then MattingAnything\cite{li2023matting} imitated Grounded-SAM to achieve matting for any object with richer details than segmentation.
Nevertheless, relying solely on human-centered dense prediction proves inadequate to meet the demands for fine-grained fashion tasks. It is necessary to investigate the integration of multiple Semantic Segmentation Models with the aim of accomplishing open-vocabulary fashion segmentation.


\subsection{Fashion Synthesis and Editing}

Previous work on human synthesis and editing usually focuses on image-to-image virtual try-on \cite{choi2021vitonhd, wang2018toward-cp,  xie2022pasta, zhu2023tryondiffusion}, or unconditional human generation \cite{fu2022stylegan}, with limited granularity and degree of control over the generation process. Recently, text-to-image fashion editing, such as Text2Human\cite{jiang2022text2human}, and HumanDiffusion\cite{zhang2022humandiffusion}, realizes the generation of human images based on pose or segmentation under the guidance of text (or labels), but these methods cannot maintain the identity of the person. FICE\cite{pernuvs2023fice} uses GAN Inversion to realize the modification of human photos based on text prompts while maintaining the characteristics of the person, but it is unable to guarantee the editing effect of images outside the distribution.
However, these methods encounter challenges in achieving meticulous control over the generated photos, or suffer from simplistic control conditions, consequently leading to functional limitations.

\section{Fashion Matrix}
\label{method}


\subsection{Overview}


The purpose of the Fashion Matrix is to enable precise and controllable fashion editing of a given photo by integrating various pre-trained models while adhering to human interaction habits. To align the framework with the collaborative workflow of a team, we have divided the Fashion Matrix into modules based on different functional roles. The pipeline of the Fashion Matrix is illustrated in Fig. \ref{fig:system}. This division not only simplifies the complexity of each module's function but also enhances the focus and efficiency of each module in its specific responsibilities. Specifically, the Fashion Matrix is divided into three modules:

\begin{itemize}
    \item \textbf{Fashion Assistant}: As a module that directly interacts with users and conveys their specific editing needs to the core editing functionality, the Fashion Assistant serves as the "customer service" or "front desk" of the framework, establishing a connection between users and the system. The Fashion Assistant primarily engages in conversations with users, collects and organizes their fashion editing requirements, and forwards them to the Fashion Designer module for further processing.
    \item \textbf{Fashion Designer}: As its name indicates, \textit{Fashion Designer} will process and optimize according to the photos to be processed and editing instructions submitted by \textit{Fashion Assistant}, and utilize BLIP\cite{li2022blip}, \textit{AutoMasker} to obtain image information, target mask, and standardized editing instructions according to the standardized processing flow. Finally, the edited image results are obtained by using various Visual Foundation Models.
    \item \textbf{AutoMasker}: It uses different human-centered parsing and pose estimation models to obtain finer-grained human semantic information and form it into CoSegmentation. Besides, AutoMasker utilizes Grounded-SAM\cite{kirillov2023segany, liu2023grounding} for open-domain segmentation to be suitable for more general fashion tasks and uses MattingAnything\cite{li2023matting} for fine-tuning of boundaries.
\end{itemize}

\subsection{Fashion Assistant}

\begin{figure}
  \centering
  \includegraphics[width=1.0\hsize]{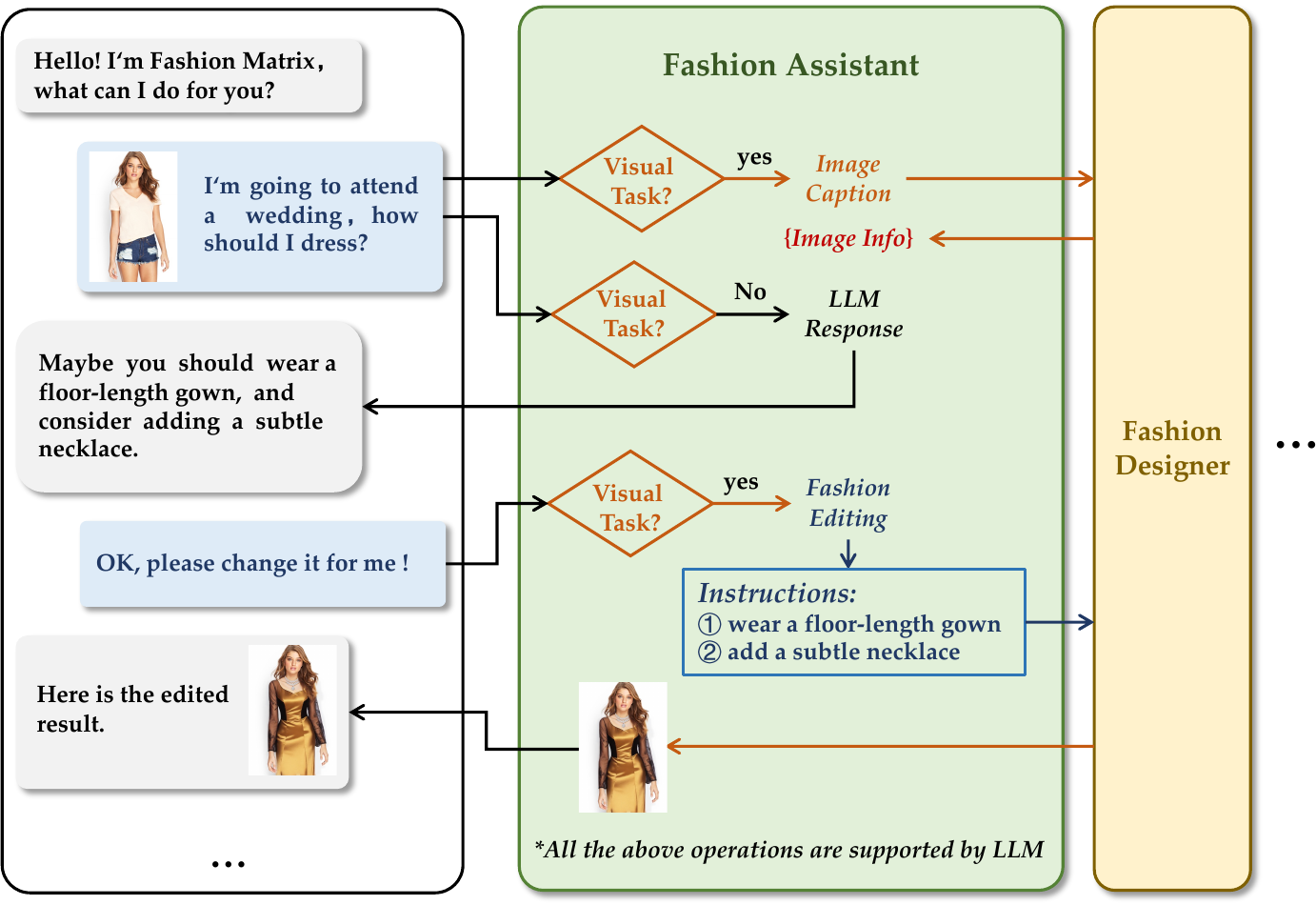}
  \caption{\textbf{The workflow of Fashion Assistant}. It possesses the capability to engage in conversations with users, maintaining context. It gathers and organizes fashion editing requirements into instructions that can be relayed to a Fashion Designer for further action.}
  \label{fig:asis}
\end{figure}

Fashion Assistant plays the role of account manager in the team, which does not directly contact the image editing business but only plays the role of docking with users and Fashion Designer. The Fashion Assistant can have natural conversations, including providing users with basic information about Fashion Matrix and answering users' questions. 

After the user uploads the image and indicates the editing instruction, the Fashion Assistant will submit the image to be edited and the editing instruction to the Fashion Designer to start the fashion editing process, then submit the editing result returned by the Fashion Designer to the user, reorganize the image to be edited according to the user's feedback on the result and submit it to the Fashion Designer according to the user's requirements, and so on. 

This clear role division and black box design avoids the confusion caused by directly letting LLMs take on too many functions, makes the processing flow clearer and clearer, and avoids designing too many systematic definitions in advance. 

\subsection{Fashion Designer}


We define Fashion Designer as a hub for receiving, processing, and distributing fashion editing tasks. The name "Designer" vividly expresses its function. 
For more controllable execution, we divide fashion tasks into 4 categories:
1) Replacement: Replace an item or partial area with another and its shape and appearance may be changed, such as modifying the neckline style.
2) Recoloring: Modify the appearance (mainly the color) of a certain part while retaining its shape, such as changing the color of pants.
3) Addition: Add an accessory or clothing that does not exist in the photo, such as add a coat, watch.
4) Removal: Erase an accessory or part, such as removing necklaces, bracelets.

After receiving the edited image $I$ and editing requirements $R$ from Fashion Assistant, Fashion Designer leverages LLM to decompose $R$ into an executable task sequence $\{T_0, ..., T_n\}$.  Each task is parameterized as $T = \{c, t_o, t_e\}$, where $c$ represents the category of the task and $t_o$ and $t_e$ represent the original and target description of the part to be edited respectively. 
$T$ and $I$ will be passed as parameters to AutoMasker to get the binary Mask $M$ used to guide image editing:

\begin{equation}
\label{eq1}
	M=AutoMasker(I,T)=AutoMasker(I,\{c,t_o,t_e\})
\end{equation}

The generation process is completed by the collaborative work of Stable Diffusion and ControlNet with different conditions. Since this process is a text-guided generation, it is necessary to obtain a suitable text prompt $t$.
For this purpose, BLIP is used to obtain more detailed information about $t_o$ in $I$, and then LLM summarizes a more appropriate text prompt from this information and $R$ for the text-to-image generation:

\begin{equation}
\label{eq2}
	t=LLM(c, BLIP(I,t_o), t_e)
\end{equation}

For recoloring, the SoftEdge version of ControlNet\cite{zhang2023Controlnet} uses the extracted edge sketch (PiDiNet\cite{su2021pidi}) to keep the shape of the target part unchanged and uses $t$ to recolor the sketch along with the Inpainting version of ControlNet. For replacement, addition, and removal, the Inpainting version of ControlNet is adopted directly:

\begin{equation}
\label{eq3}
	I_e=\left\{  
             \begin{array}{lr}  
             G(I,M, PiDi(I),t),&  if\; c \;is\; recoloring\\  
             G(I,M,t),&  else\\  
             \end{array}  
\right.  
\end{equation}
where $\bigodot$ is the element-wise multiplication, $G(\Theta)$ represents the Stable Diffusion\cite{rombach2021highresolution} generation process controlled by ControlNets\cite{zhang2023Controlnet}.

\subsection{AutoMasker}


To make use of fine-grained mask semantics and control mask generation, we propose an AutoMask module to balance the degree of original information retention and the naturalness of generation fusion. 

For an input image $I$, AutoMasker first processes the human semantic segmentation (Graphonomy) and pose estimation (DensePose), and combine them to obtain a more fine-grained semantic segmentation map named CoSegmentation $S_{co}=\{s_0:m_0,…,s_n:m_n \}$, where $s_i$ represents the semantic of certain part, and $m_i$ represents the mask corresponding to the semantic. Despite the absence of certain common semantics in the original image, they can be effectively estimated by employing various operations such as combination, cropping, and pooling, based on the existing semantics. This, in turn, greatly facilitates the Addition task. The visualization of CoSegmentation is as Fig.\ref{fig:cosegm}:

\begin{figure}
  \centering
  \includegraphics[width=1\hsize]{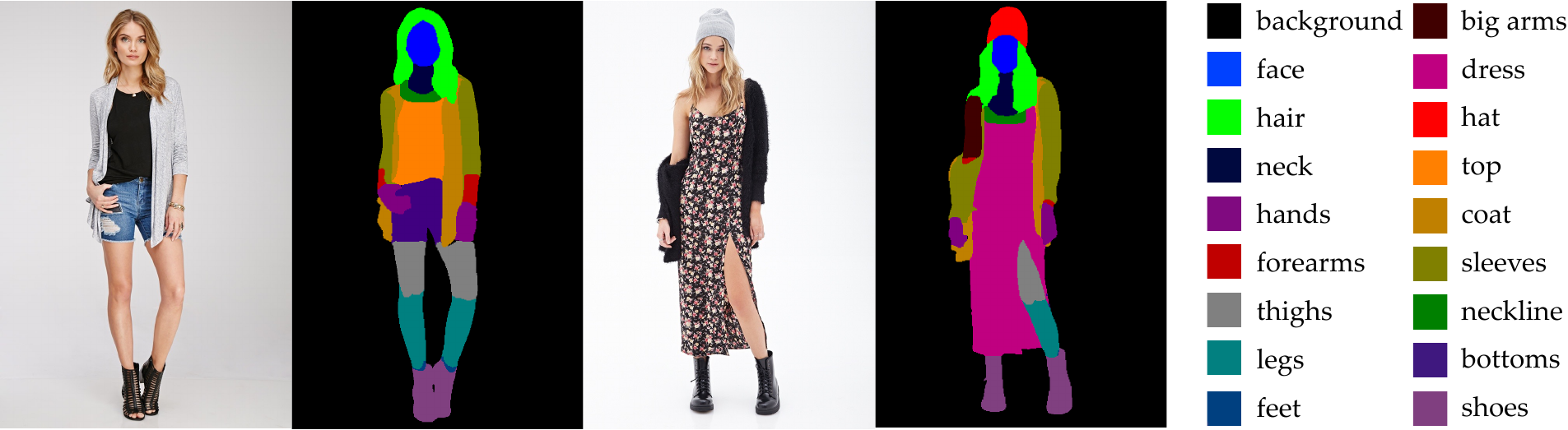}
  \caption{\textbf{Visualization of CoSegmentation} which combines Graphonomy\cite{Gong2019Graphonomy} with DensePose\cite{Guler2018DensePose} to obtain a more fine-grained semantic segmentation map.}
  \label{fig:cosegm}
\end{figure}

To make full use of the task information $T=\{c,t_o,t_e\}$ from Fashion Designer, AutoMasker adopts different mask schemes according to the task category $c$.

For recoloring, replacement, and removal tasks, AutoMasker utilizes LLM to judge if the semantic $s_i$ corresponding to the original part $t_o$ in $S_{co}$. If in, $m_i$ is adopted as the original part mask $m_o$ directly. If not, AutoMasker will utilize GroundingSAM to obtain the original part mask $m_o$ from $I$ and $t_i$.Therefore $m_o$ can be obtained by the following formula:

\begin{equation}
\label{eq4}
	m_o=\left\{  
             \begin{array}{lr}  
             S_{co}[s_i]=   S_{co}[LLM(t_o)],&  if\; s_i \;in\;  S_{co}\\  
             GroundingSAM(t_o),&  else\\  
             \end{array}  
\right.  
\end{equation}

For the removal task, there is no need to consider the target object or part, so its final mask can be obtained from:

\begin{equation}
\label{eq5}
    M_{remove}=MaxPool(m_o)
\end{equation}

where $MaxPool$ represents the boundary expansion of Mask by using maximum pooling, which can make local editing more consistent and coordinated with the surrounding context.

For the recoloring task, it is necessary to ensure that only the target area is recolored. Therefore, MAM is additionally used to alleviate the problem of $m_o$ boundary blur and possible overlap at the junction of the background. The process can be expressed as:

\begin{equation}
\label{eq6}
    M_{recolor}=MAM(I) \bigodot m_o
\end{equation}

For the replacement task, besides $m_o$, other parts may also be occluded by the target object. For instance, when replacing a vest with a t-shirt, part of the arms may also be obstructed.  To address this issue, AutoMasker uses LLM to logically infer the body parts that may be masked by the target object $S_m=\{m_0^m,…,m_k^m\}$ from $S_{co}$ which are merged with $m_o$ to form a more reasonable target mask:
\begin{equation}
\label{eq7}
    S_m=LLM(S_{co}, t_e)=\{m_0^m,…,m_k^m\}
\end{equation}

\begin{equation}
\label{eq8}
    M_{replace}=MaxPool(m_o+\sum_{m_i^m\in S_m}m_i^m)
\end{equation}

For the addition task, the target object or area does not exist in $S_co$, so it is only necessary to infer the parts that the target object may mask:
\begin{equation}
\label{eq9}
    M_{add}=MaxPool(\sum_{m_i^m\in S_m}m_i^m)
\end{equation}


\section{Experiment}
\subsection{Implementation Details}

\textbf{Visual Foundation Models.} We choose Realistic Vision V4.0 finetuned from Stable Diffusion V1.5\cite{rombach2021highresolution} as the base generator. This model can largely alleviate the unrealistic problems of characters' faces and hands. The SoftEdge and Inpainting variants of ControlNet v1.1\cite{zhang2023Controlnet} are employed for conditional control purposes. We employ BLIP\cite{li2022blip} for visual question answering.

\textbf{Semantic Segmentation Models.} To obtain human-centric dense predictions, we first employ Graphonomy\cite{Gong2019Graphonomy} and DensePose\cite{Guler2018DensePose}. Subsequently, we utilize Grounded-SAM\cite{kirillov2023segany, liu2023grounding} and MattingAnything\cite{li2023matting} to facilitate open-vocabulary segmentation acquisition and edge refinement.

\textbf{Large Language Models.} As the logical reasoning and dialogue with users of the system is supported by LLMs, we have conducted a series of evaluations on various open-source LLMs configured with distinct parameter level, encompassing FastChat-T5-3B\cite{zheng2023judging}, ChatGLM-6B\cite{zeng2022glm}, ChatGLM2-6B\cite{zeng2022glm}, Vicuna-7B\cite{zheng2023judging}, Vicuna-13B\cite{zheng2023judging}, and Baichuan-13B-Chat\cite{Baichuan-13B}.



\subsection{Comparison with Text-to-Image Baselines}
Currently, there is a lack of a text-based fashion editing method for images. In this regard, we compare our system with two existing text-based try-on approaches: Text2Human\cite{jiang2022text2human} and FICE\cite{pernuvs2023fice}. Text2Human is capable of generating try-on results based on the pose and parsing conditions. However, it should be noted that text-based try-on constitutes only a minor component of the functionalities offered by Fashion Matrix.
In our comparative analysis, we observed that Fashion Matrix outperformed these two methods in terms of CLIP Score\cite{radford2021clipscore} and IS\cite{salimans2016improved_is}.
Furthermore, we conducted a human evaluation to assess the naturalness and text-image matching of the generated images. For each criterion, we randomly assembled 30 sets of result images, and volunteers were tasked with selecting the option that appeared more natural or exhibited superior text-image matching. We obtained 25 responses for assessing naturalness and 20 responses for evaluating text-image matching.
The evaluation showed in Table \ref{tab:quant_comparison} demonstrates that Fashion Matrix produces outputs with improved realism, naturalness, and adherence to text descriptions.



\begin{table}[]
    \centering
    \caption{\textbf{Quantitative comparison} with Text2Human\cite{jiang2022text2human} and FICE\cite{pernuvs2023fice}.In addition to assessing the CLIP Score\cite{radford2021clipscore} and Inception Score (IS)\cite{salimans2016improved_is}, we conducted evaluations on the naturalness and text-image matching. Our system possess advantages across all these metrics.}
    \vspace{3mm}
    \begin{tabular}{c|cccc}
        \toprule
        \multirow{2}{*}{Method} & \multicolumn{1}{c}{\multirow{2}{*}{CLIP Score$\uparrow$}} & \multicolumn{1}{c}{\multirow{2}{*}{IS$\uparrow$}} & \multicolumn{2}{c}{Human Evalution} \\ \cline{4-5} 
                                  & \multicolumn{1}{c}{} & \multicolumn{1}{c}{} & Naturalness & Text-Image Matching \\ \midrule
        FICE                      & 23.74                & 2.54                 & -           &   -                  \\
        Text2Human (from parsing) & 26.49                & 3.10                 & 23.33\%     &  28.13\%             \\
        Text2Human (from pose)    & 26.63                & 3.04                 & 21.33\%     &  25.20\%              \\
        Ours                      & \textbf{27.78 }      & \textbf{3.14}        & \textbf{55.33\%}     & \textbf{ 46.67\%}          \\
        \bottomrule
    \end{tabular}
    \label{tab:quant_comparison}
\end{table}

\subsection{Ablation Studies}
In our study, we employed ChatGPT\cite{InstructGPT} to generate requirements with limited examples from users. Subsequently, we classified these instances into three distinct categories: single-task, dual-task, and multi-task requirements, comprising 100, 70, and 50 instances, respectively. In task classification, we exclusively employ the set of 100 single-task requirements. We proceeded to manually rectify the task splitting and classification results generated by LLMs, utilizing these cases as a benchmark for evaluating the efficacy of various LLMs in handling fashion-related tasks.
We conduct evaluations on the 6 aforementioned open-source LLMs without individually optimizing prompts for each model. Nevertheless, it is essential to acknowledge that the stochastic nature of LLMs and the inherent bias towards specific prompts may lead to the fact that our evaluation results do not fully reflect the capabilities of LLMs.

As shown in Table \ref{ablation-table}, for task classification, there exists a positive correlation between the accuracy rate and the number of model parameters. Specifically, both the 13B models attained an accuracy rate exceeding 75\%, whereas the performance of the remaining models was comparatively lackluster.
However, in the case of task splitting, it is noteworthy that the accuracy rate does not exhibit a significant correlation with the number of model parameters. For instance, despite having relatively less parameters, Vicuna-7B\cite{zheng2023judging} and FastChat-T5-3B\cite{zheng2023judging} demonstrated impressive performance. Conversely, Baichuan-13B-Chat\cite{Baichuan-13B} struggled to produce accurate results.
Overall, considering all the factors, Vicuna-13B\cite{zheng2023judging} emerges as the most suitable option for supporting Fashion Matrix.

\begin{table}
  \caption{\textbf{Accuracy comparison} of different LLMs for Task Splitting and Classification.For Task Splitting, we divide the test cases into single-task, dual-task, and multi-task requirements, which are represented by 1, 2 and 3 + in the table respectively.}
  \label{ablation-table}
  \centering
  \begin{tabular}{c|ccccc}
    \toprule
    \multirow{2}{*}{LLM} & \multicolumn{4}{c}{Task Splitting} & \multicolumn{1}{c}{\multirow{2}{*}{Task Classification}}\\ 
    \cline{2-5} 
    \multicolumn{1}{c|}{} & 1 & 2 & 3+ & Average & \multicolumn{1}{c}{}  \\ 
    \midrule
    Vicuna-13B\cite{zheng2023judging}   & 73.00\%       & 87.14\%     & 78.00\%         & 78.64\%     &\textbf{78.00\%}  \\
    Baichuan-13B-Chat\cite{Baichuan-13B}& 10.00\%       & 64.29\%     & 10.00\%         & 27.27\%     &75.00\%  \\
    Vicuna-7B\cite{zheng2023judging}    & 86.00\%       & \textbf{94.29\%} & \textbf{88.00\%} & \textbf{89.09\%}    &45.00\%  \\
    ChatGLM-6B\cite{zeng2022glm}        & 70.00\%       & 81.43\%     & 78.00\%         & 75.45\%     &60.00\%  \\
    ChatGLM2-6B\cite{zeng2022glm}       & 75.00\%       & 65.71\%     & 62.00\%         & 69.09\%     &42.00\%  \\
    FastChat-T5-3B\cite{zheng2023judging}& \textbf{87.00\%} & 81.43\%     & 86.00\%         & 85.00\%     &53.00\%  \\
    \bottomrule
  \end{tabular}
\end{table}

\section{Conclusion}
In this work, we propose Fashion Matrix, a multi-round dialogue AI system that integrates LLM, Visual Foundation Models, and Semantic Segmentation Models to realize text-guided fashion editing tasks with open vocabulary. 
This system innovatively integrates various Semantic Segmentation Models to construct a  more detailed semantic map called CoSegmentation and adaptively generates task-specific masks for the editing area, which effectively address complex text-guided fashion editing tasks.
Extensive experiment and specific cases have demonstrated the remarkable potential and proficiency of Fashion Matrix in various fashion editing tasks.
Nevertheless, the optimization of LLMs specifically for the fashion domain, alongside the development of detailed Semantic Segmentation Models for both human subjects and fashion items, harbors considerable potential in elevating and broadening the functionalities of the Fashion Matrix.


\newpage

\begin{figure}
  \centering
  \includegraphics[width=1.0\hsize]{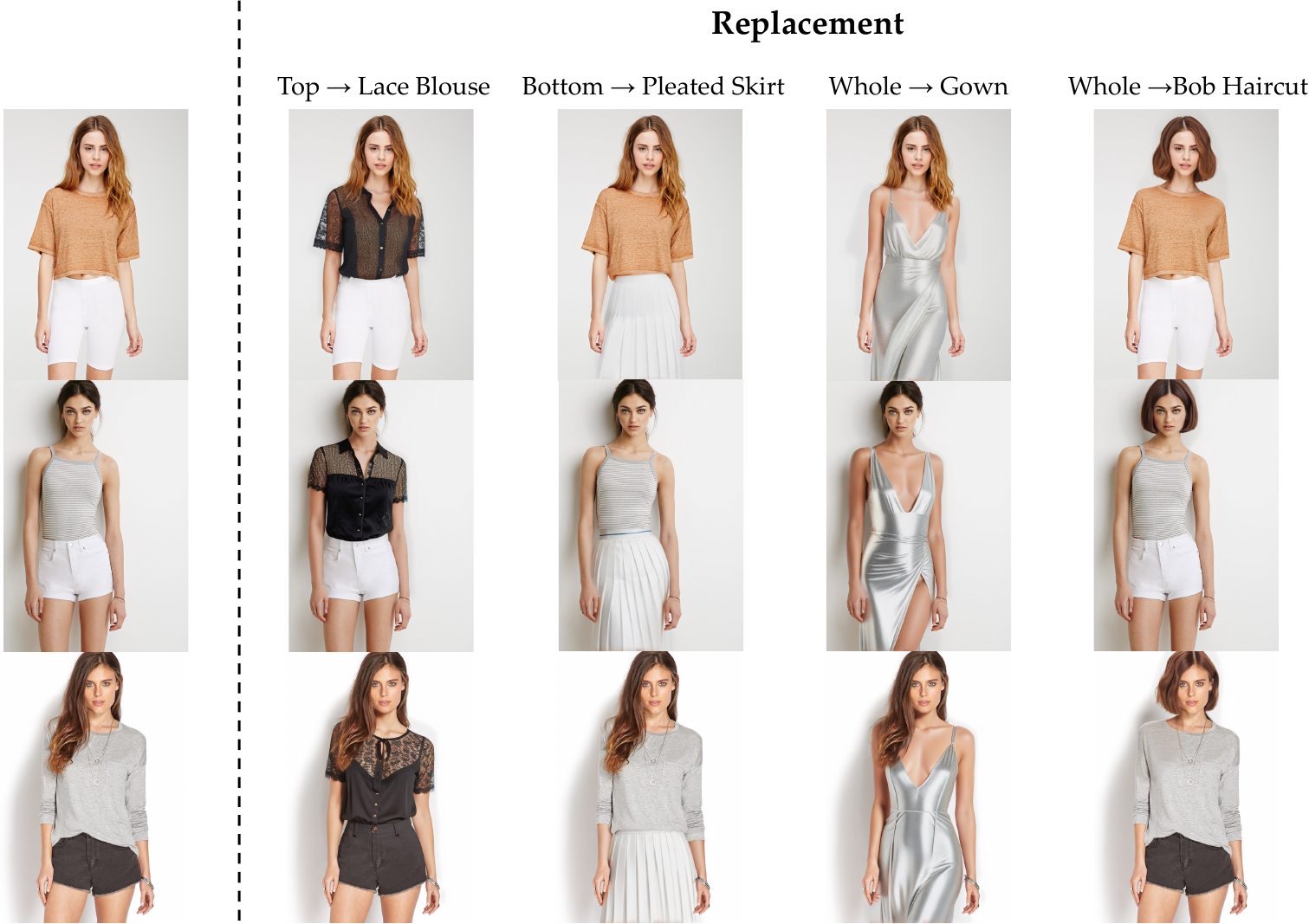}
  \caption{\textbf{Results of the replacement task} implemented by Fashion Matrix. The replaced target is integrated with the source image seamlessly, adeptly accounting for authentic lighting conditions and occlusion challenges.}
  \label{fig:replace}
\end{figure}

\begin{figure}
  \centering
  \includegraphics[width=1.0\hsize]{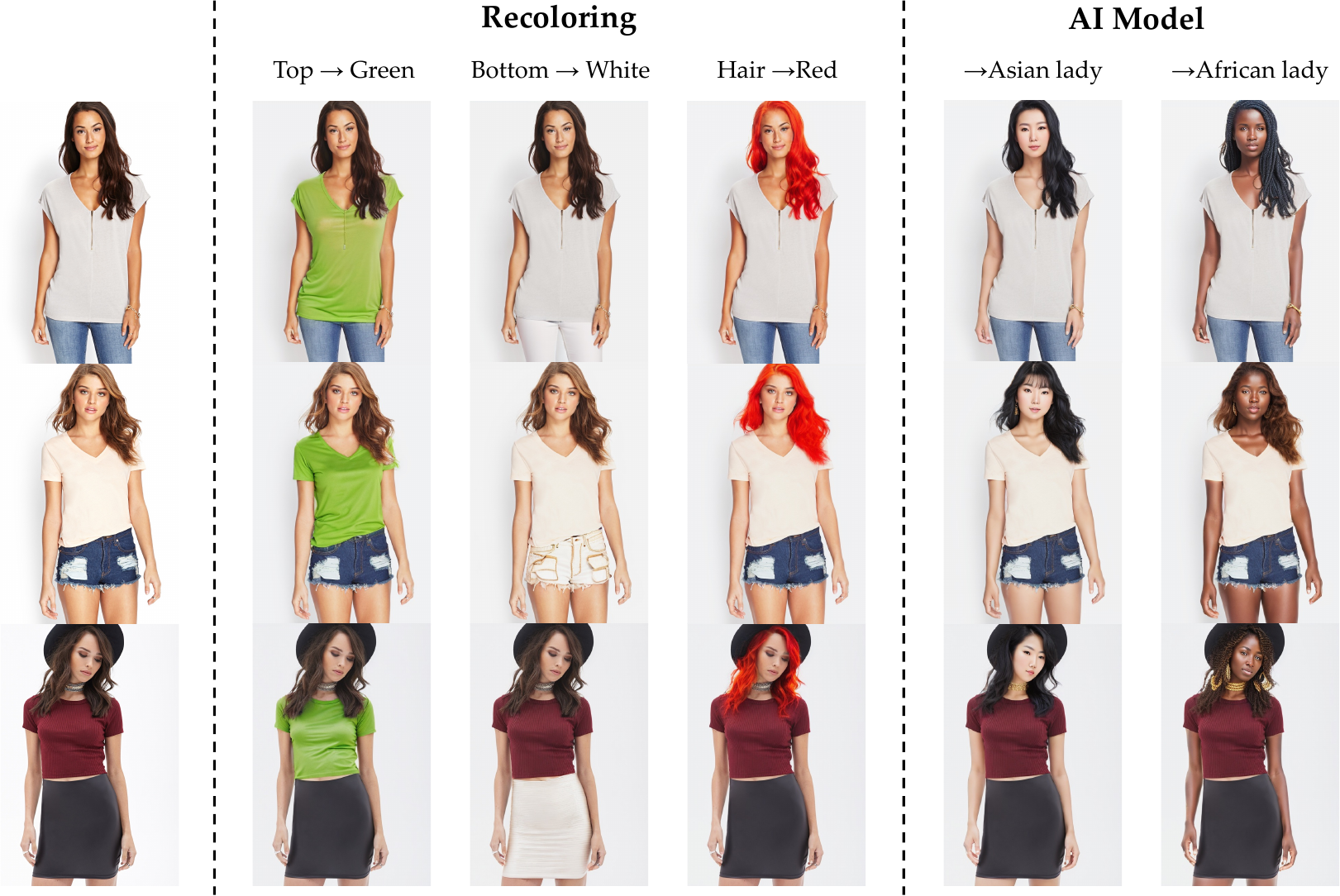}
  \caption{\textbf{Results of the recoloring task} implemented by Fashion Matrix. Utilizing a dual set of mask and edge sketch, the recoloring process is conjointly regulated, ensuring seamless integration of the generated output with the unaltered regions, while preserving the shape of the original entity.}
  \label{fig:recolor}
\end{figure}

\begin{figure}
  \centering
  \includegraphics[width=0.9\hsize]{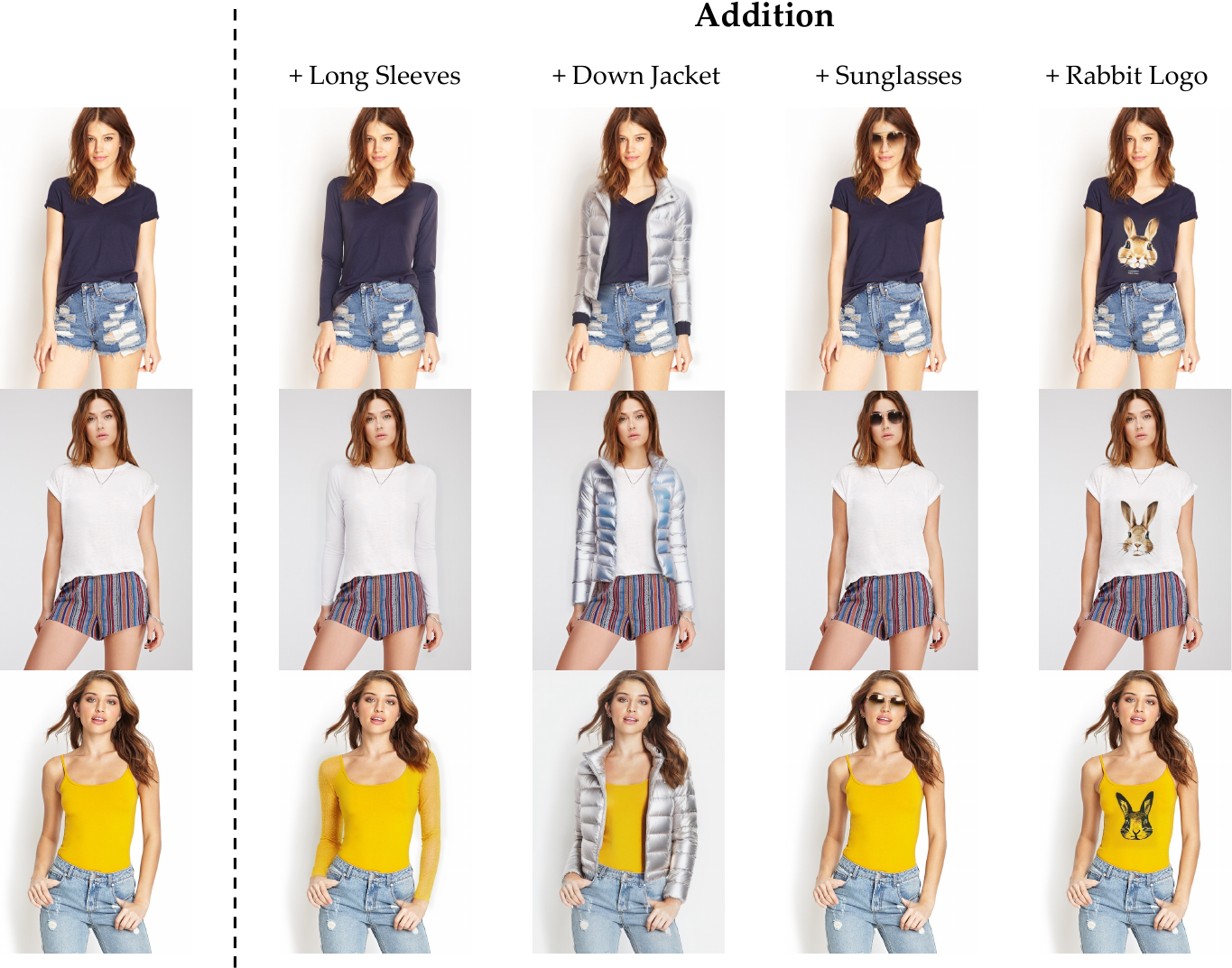}
  \caption{\textbf{Results of the replacement task} implemented by Fashion Matrix. Based on the positioning capabilities of CoSegmentation and LLM, Fashion Matrix facilitates the incorporation of non-existent items into an image, while ensuring coherence between the newly added entity and the original visual context.}
  \label{fig:add}
\end{figure}

\begin{figure}
  \centering
  \includegraphics[width=0.96\hsize]{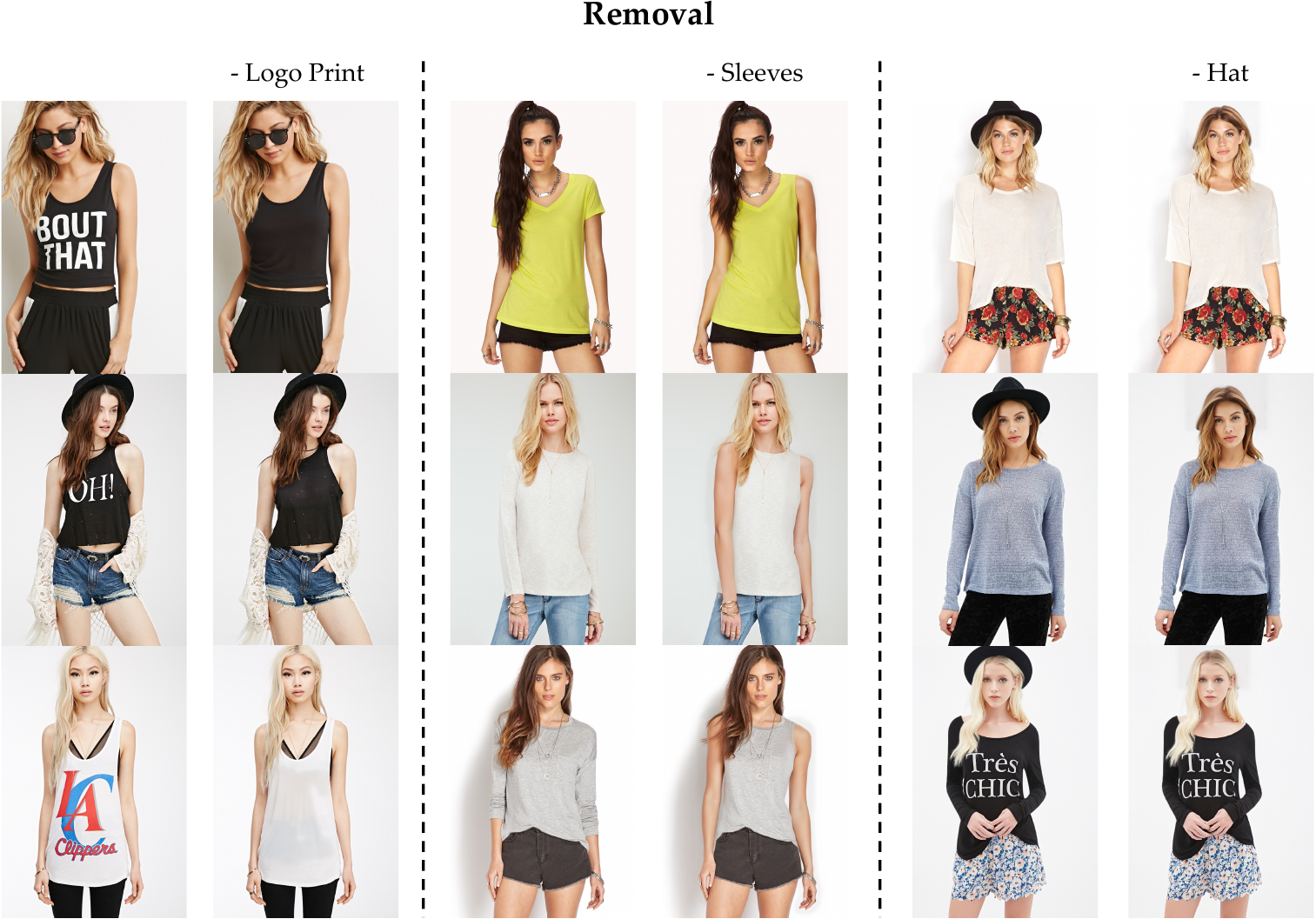}
  \caption{\textbf{Results of the replacement task} implemented by Fashion Matrix. Fashion Matrix's identification of entities for removal is informed by the fine-grained CoSegmentation, coupled with the open domain segmentation capabilities offered by Grounded-SAM\cite{kirillov2023segany, liu2023grounding}.}
  \label{fig:remove}
\end{figure}

\end{document}